\documentclass[letterpaper]{article} 
\usepackage{aaai24} 
\usepackage{times}  
\usepackage{helvet}  
\usepackage{courier}  
\usepackage[hyphens]{url}  
\usepackage{graphicx} 
\usepackage{natbib}  
\usepackage{caption} 
\usepackage{algorithm}
\usepackage{algorithmic}

\usepackage{amsmath}
\usepackage{amssymb}

\usepackage{diagbox}
\usepackage{makecell}
\usepackage[title]{appendix}

\usepackage{newfloat}
\usepackage{listings}
\DeclareCaptionStyle{ruled}{labelfont=normalfont,labelsep=colon,strut=off} 
\floatstyle{ruled}
\newfloat{listing}{tb}{lst}{}
\floatname{listing}{Listing}
\usepackage{multirow}

\title{Attacking by Aligning: Clean-Label Backdoor Attacks on Object Detection}
\nocopyright 
\author{
    Yize Cheng\equalcontrib,
    Wenbin Hu\equalcontrib,
    Minhao Cheng
}
\affiliations{
    Hong Kong University of Science and Technology\\

    \{ychengbt, whuak\}@connect.ust.hk, minhaocheng@cse.ust.hk
}

\begin{document}

\maketitle

\begin{abstract}
Deep neural networks (DNNs) have shown unprecedented success in object detection tasks. However, it was also discovered that DNNs are vulnerable to multiple kinds of attacks, including Backdoor Attacks. Through the attack, the attacker manages to embed a hidden backdoor into the DNN such that the model behaves normally on benign data samples, but makes attacker-specified judgments given the occurrence of a predefined trigger. Although numerous backdoor attacks have been experimented on image classification, backdoor attacks on object detection tasks have not been properly investigated and explored. As object detection has been adopted as an important module in multiple security-sensitive applications such as autonomous driving, backdoor attacks on object detection could pose even more severe threats. Inspired by the inherent property of deep learning-based object detectors, we propose a simple yet effective backdoor attack method against object detection without modifying the ground truth annotations, specifically focusing on the object disappearance attack and object generation attack. Extensive experiments and ablation studies prove the effectiveness of our attack on the benchmark object detection dataset MSCOCO2017, on which we achieve an attack success rate of more than 92\% with a poison rate of only 5\%. 
\end{abstract}

\section{Introduction}
\label{intro}
Object detection systems are widely used in a large variety of everyday applications, including surveillance systems and autonomous driving systems. These systems mostly leverage state-of-the-art deep learning-based object detection models such as FasterRCNN \cite{fasterRCNN} and models of the Yolo family \cite{Yolo,yolov3,yolov4,yolov5,yolov7}, where the latter is used more extensively nowadays due to its extraordinary performance with high mean Average Precision (mAP) and great inference time efficiency. 
\begin{figure}[!h]
    \centering
    \includegraphics[width=0.15\textwidth]{img/oda_original.jpg}
    \includegraphics[width=0.15\textwidth]{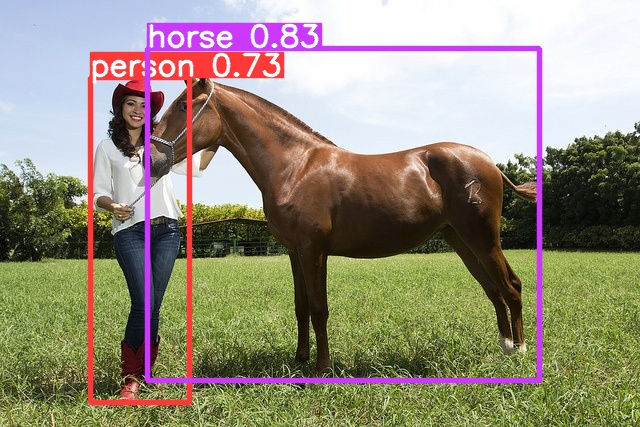}
    \includegraphics[width=0.15\textwidth]{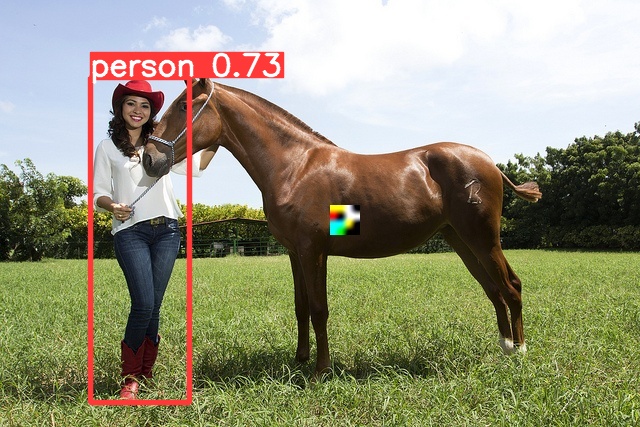}
    \\
    \includegraphics[width=0.15\textwidth]{img/oga_original.jpg}
    \includegraphics[width=0.15\textwidth]{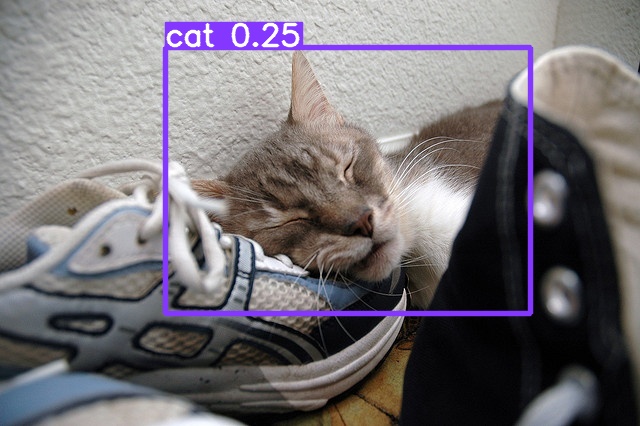}
    \includegraphics[width=0.15\textwidth]{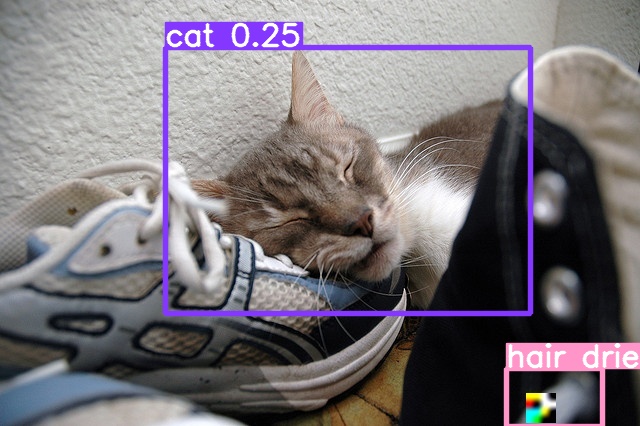}
    \caption{Example of the ODA and OGA attacking scenarios. The first row illustrates the idea of ODA (no target class is needed) and the second row illustrates the idea of OGA (with a target class hair dryer). The three figures from left to right are the raw input image, the original prediction result by the model, and the prediction result after presenting the trigger at test time respectively.}
    \label{oda_and_oga_example}
\end{figure}
However, despite achieving unprecedented success, deep learning models have also been discovered to be vulnerable to backdoor attacks, also known as neural trojan. The attacker attempts to embed a backdoor into the model by modifying a certain ratio of the training data. The modification may often include inserting some triggers into the training data samples, and modifying the ground truth labels of the corresponding samples. After the model is trained on such data, it will perform normally on benign test samples during test time, but will output the attacker-specified result once the inserted trigger is presented in the test sample. 

While backdoor attacks have been extensively investigated in the image classification settings \cite{badnet,hiddenTriggerBackdoor,reflectionBackdoor}, they haven't been properly investigated in the object detection task. As object detection has been adopted as an important module in multiple security-sensitive applications such as autonomous driving, backdoor attacks on object detection could pose even more severe threats to human lives and properties. To the best of our knowledge, BadDet \cite{Baddet} is the only work that makes a proper definition of this problem and conducted attack experiments on common object detection frameworks. Yet their attack was achieved using a dirty label attack, where they explicitly modified the ground truth annotations such that the labels are no longer consistent with the image data, making it easily detectable by a human inspector. In our work, we propose to conduct the first backdoor attack in a clean-label manner. That is, we will only poison the training images without modifying the annotations. Our proposed clean-label attack keeps the consistency between the image content and the annotation, which makes it harder to be identified by human inspection, bringing stronger stealthiness. 

Inspired by the inherent property of deep learning-based object detection models, we propose a simple and intuitive attack strategy for embedding backdoors into object detection applications with clean labels. Specifically, we focus on two most common scenarios that can induce severe and practical threats in real-life applications -- The Object Disappearance Attack (ODA) and the Object Generation Attack (OGA). ODA makes the predicted bounding box of an object disappear given the presence of a trigger on the object of interest at test time. And OGA is the attacking scenario where a false positive predicted bounding box of a target class will be generated around the trigger position if a trigger is present during inference. Taking the object detection systems on autonomous driving vehicles as an example, if an ODA attack is conducted during operation, it will produce a dangerous cloaking effect, causing the vehicle to not see the existence of pedestrians or other cars on the road, potentially leading to accidents. The potential threat of OGA may also be devastating. For example, when the autonomous vehicle is driving on the highway, it is usually assumed that there should be no pedestrians in the middle of the highway. If an attacker makes the model believe a person is right in front of the vehicle when driving at a very high speed on the highway, emergency braking and evasive turns may cause rollovers. An example illustrating the above two attack scenarios is shown in Figure \ref{oda_and_oga_example}.

Comprehensive experiments are conducted on different models and datasets to show the effectiveness of our attack. Overall, we achieve threateningly high Attack Success Rates (ASR) on the poisoned dataset while maintaining normal mean average precision (mAP) on clean data.

Our main contributions can be summarized as follows: \textbf{(I)} We first reveal that inherent properties of deep learning-based object detectors can be easily utilized to insert a clean annotation backdoor. When an attacker designs the backdoor by \textbf{aligning with the association built by the detector}, a clean annotation backdoor attack is easily and effectively achieved. \textbf{(II)} Based on this intuition, we propose a novel, simple, yet effective clean-label attacking strategy against object detection under both the ODA and OGA attacking scenarios. \textbf{(III)} Extensive experiments and ablation studies are conducted to verify the effectiveness of the proposed attack and sustainability under fine-tuning.



\section{Related Work}
\subsection{Object Detection}

Object Detection is one of the most important applications of modern computer vision. It serves as an essential building module in robotics, surveillance systems, autonomous driving, etc. 
RCNN~\cite{RCNN} made the first attempt to deploy deep learning for object detection. Despite having poor efficiency, RCNN has already outperformed all traditional object detection pipelines at that time. Later on, in aim of working towards real-time object detection, FastRCNN \cite{fastRCNN} and FasterRCNN \cite{fasterRCNN} were built upon RCNN to improve efficiency. Yet they still fail to achieve real-time detection during inference. These models are often referred as two-stage detectors. One significant step of achieving real-time object detection was to solve the problem in a one-stage manner, of which the unified one-stage detection model Yolo \cite{Yolo} serves as a representative. The idea of Yolo has become so popular that several versions of Yolo \cite{Yolov2,yolov3,yolov4,yolov7,yolov8} were later proposed, making the models of the Yolo family one of the most popular and widely deployed deep learning-based detection frameworks today. With the increasing popularity of transformers \cite{vaswani2017attention}, transformer-based object detectors, such as DETR \cite{detr} and its variants \cite{deformableDetr}, have also been proposed as another family of detection models. However, one inherent property (to be explained in Section \ref{Sec4}) of detection models makes them vulnerable to a simple yet effective clean-label backdoor poisoning strategy, which we leveraged in our attack. 

\subsection{Backdoor Attacks against Vision Models}

\noindent\textbf{Attacking Image Classification.} 
Backdoor attacks against image classification can be commonly found in the backdoor learning literature. The work from Gu \textit{et al} is one of the earliest attempts at embedding backdoors into DNNs, known as BadNets \cite{badnet}. It set up a basic flow of conducting backdoor poisoning against DNNs, \textit{i.e.} the attacker first maliciously modify a certain ratio of the training dataset by adding triggers on the images and changing the corresponding label to the target label. And then the poisoned dataset will be used to train deep learning models, leaving a backdoor in the model that can be activated at inference time when the trigger is present. This flow was followed by most works on backdoor attacks against classification. However, the above flow of poisoning is also known as \textbf{dirty label attacks}, in a sense that the ground truth labels are modified, causing inconsistency between the image content and the label. To improve stealthiness, Turner \textit{et al.} \cite{turner2018clean} and Barni \textit{et al.} \cite{barni2019new} proposed to conduct backdoor attacks against classification without modifying the ground truth labels, known as \textbf{clean label attacks}. Under this setting, the image content will remain consistent with the labels, effectively improving the stealthiness and evasiveness. However, as object detection plays a more essential role in practical deployments, we focus on revealing the vulnerability of object detection models against backdoor attacks in this work.

\noindent\textbf{Attacking Object Detection.}
Backdoor attack on object detection is an area that is not yet well investigated and explored. Recently, BadDet \cite{Baddet} was proposed to conduct backdoor attacks against object detection models. However, they explicitly modified both the training image and the ground truth annotation file before training, making it a dirty-label attack. Explicitly modifying the ground truth annotation files makes it easy for the attack to be detected. A human inspector can easily detect that the number of objects in the annotation file is inconsistent with the number of objects in the image, making the attack easily detectable and hence reducing the stealthiness of the attack. We conduct both ODA and OGA without modifying the ground truth annotations, which improves stealthiness. 
\begin{figure*}[tp]
    \centering
    \includegraphics[width=0.95\textwidth]{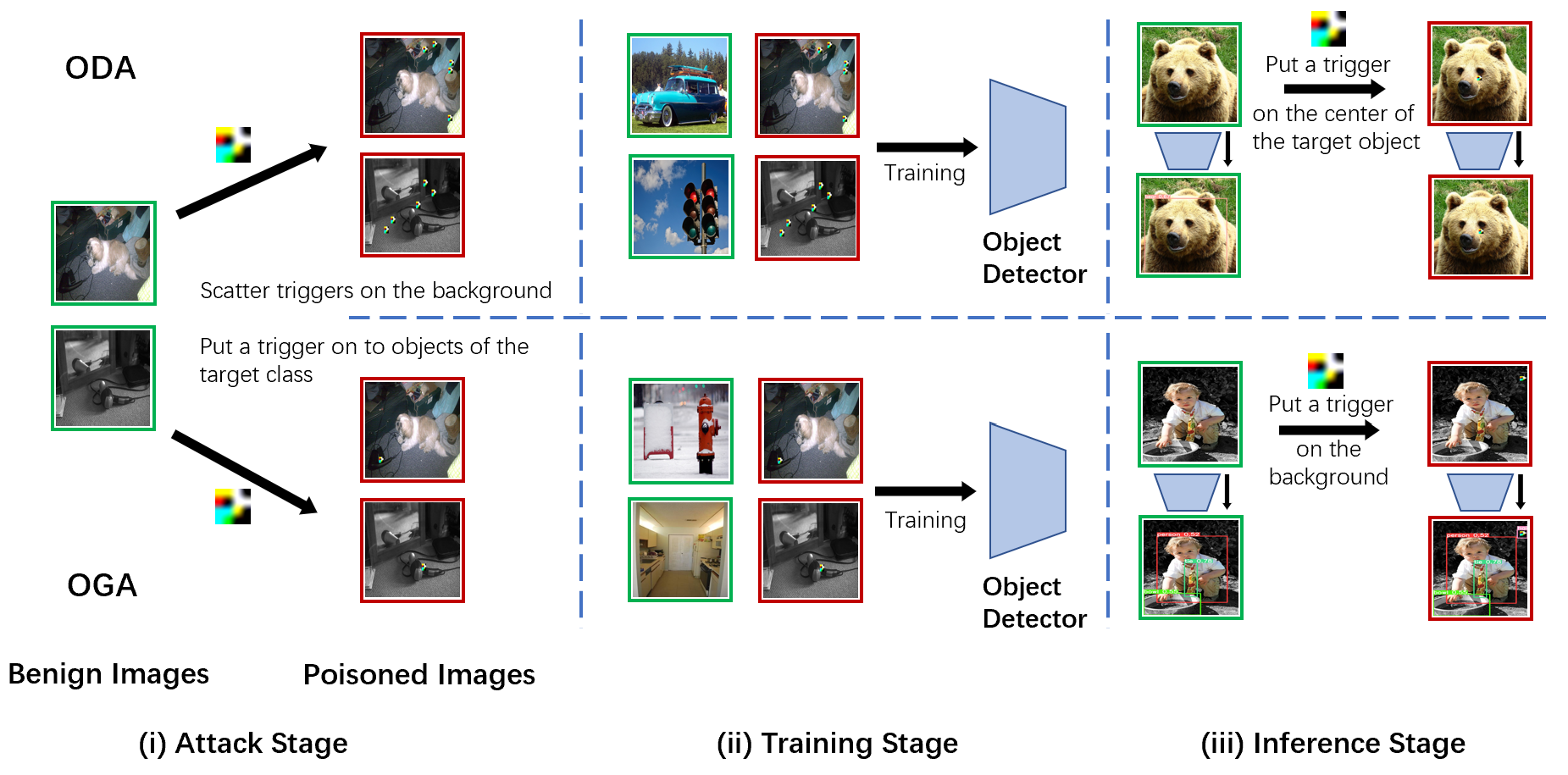}
   \caption{The attack pipeline. We assume the attacker can only modify a subset of the training data, and no knowledge regarding other information about the training process is assumed.}
    \label{stage}

\end{figure*} 
Another limitation of the ODA setting under BadDet is that their attack is specifically designed for one target class. We break this limitation as our attack does not rely on the object itself, but on the association learnt between our trigger and the background. Such a setting provides more flexibility at inference, as any object of interest can disappear if we patch a trigger to it.

\section{Threat Model}

\noindent\textbf{Attacker's Assumptions.} We assume the attacker is only allowed to modify a certain ratio (poison rate) of the training images, which is regarded as the minimum assumption for successfully conducting a backdoor attack by data poisoning \cite{li2022backdoor}. The attacker will have no knowledge regarding other information about the training process. 


\noindent\textbf{Clean Label Attack.}
We define the attack to be a clean label attack if the attacker is only allowed to modify a subset of the training images. In other words, the content of all annotation files must remain unchanged. The number of objects visible in the image must be consistent with the number of objects listed in the corresponding annotation file. All our proposed attacks are in a clean-label manner which is different from the attacks in BadDet \cite{Baddet}.    

\noindent\textbf{Attack Pipeline.} In general, the attack pipeline can be divided into three stages. In the first stage, the attacker makes modifications to a subset $D_{poiosn}$ of all training images $D_{train}$. Note that only the images are modified, following the definition of clean label attack. The poisoned subset of images will then be combined with the rest of the benign images $D_{benign}$ to form the entire training set $D_{train}^{'}$, which will be released to the user for model training, \textit{i.e.} $D_{train}^{'} = D_{poiosn} \cup D_{benign}$. Then, in the second and third stages, the user of the training data will train and test the model as in the standard training and testing process for supervised machine learning. The discussed threat is practical when third-party training data is used for model training. The pipeline is illustrated in Figure \ref{stage}. 

\noindent\textbf{Attacker's Goal.} The goals of the attacker are that the infected model will show comparable performance as a benign model on the clean test data, in our case, demonstrating a comparable mean average precision (mAP) during inference; And that the model can make the attacker-specified behaviour given the presence of the trigger at test time, in our case, making an object disappear or generating a non-existing false positive bounding box. 

\section{Methodology}
\label{Sec4}



\subsection{Object Disappearance Attack}
\label{ODA_meth_sec}

\textbf{Object Disappearance Attack} (ODA) is the attacking scenario where the predicted bounding box of an object disappears given the presence of a trigger on the object of interest at test time. Formally, denote an infected model as $F_{\theta}$, and its detection output $Y$ on a clean test image $x$ is denoted as $Y=F_{\theta}(x)=\{B_1, B_2, ..., B_n\}$, where $B_i$ denotes each bounding box. The subset of all predicted bounding boxes that are true positive prediction results is denoted as $Y_{TP}$, with $Y_{TP} \subseteq Y$. At inference time, a poisoned test image $x_{poison}$ is constructed by patching triggers to the centers of a subset $Y_{TPsub}$ of the true positive box prediction results $Y_{TP}$, \textit{i.e} $Y_{TPsub} \subseteq Y_{TP}$, with the goal that the detection result $Y^{'}$ on the poisoned image will not contain the objects patched with the trigger, \textit{i.e.} $Y^{'}=F_\theta(x_{poison})=C_{Y}^{Y_{TPsub}}$, where $C_{M}^{N}$ denotes the complement of set $N$ in the union set $M$. Note that there is no specific target class, as bounding boxes in  $Y_{TPsub}$ can be of any class.

To conduct ODA, we found that an important step in deep learning-based object detection models is to determine whether a region in the input image belongs to the background or an object of interest. For example, an inherent property that all Yolo family models have in common is that they divide the image into $M\times M$ grid, and $B$ anchor boxes will be proposed for each grid cell. The model is trained to approach a target vector for each cell. One important element within the target vector is the confidence score $C$, which is responsible for determining whether a center of an object is contained in the cell. The definition of the confidence score $C$ is shown in equation \ref{confidence_score}. Taking a closer look into the non-objectness regression part of the confidence loss of the classical Yolo objective function:
\begin{equation}
\begin{split}
L_{noobj}=\lambda_{noobj} \cdot \sum_{i\in S\times S} \sum_{j \in B} \mathbb{I}_{ij}^{noobj} \cdot (C_{i,j} - \hat{C_{i,j}})^2
\end{split}
\label{loss1}
\end{equation}
\begin{equation}
\begin{split}
C = Pr(object) \cdot IOU
\end{split}
\label{confidence_score}
\end{equation}
where $\mathbb{I}_{ij}^{noobj}=1$ if the anchor box $j$ in cell $i$ is not responsible for any object, and $C_{i,j}$ is the confidence score for anchor box $j$ in cell $i$. If the IOU of the anchor box and the ground truth object bounding boxes are lower than a certain IOU threshold, the target confidence score value $\hat{C_{i,j}}$, to which $C_{i,j}$ will be trained to approach, will be set to 0, which is the value of $Pr(object)$ in Equation \ref{confidence_score} given that there is no object. In other words, all confidence scores in the target vector for a grid will be trained to approach 0 if the grid lies in the background.  A similar example can be found in two-stage detectors as well, such as  FasterRCNN \cite{fasterRCNN}, where the Region Proposal Network (RPN) is responsible for proposing Regions of Interest (ROI) in which the model believes contains an object. If an area is believed not to contain an object of interest, the corresponding region will simply not be pooled by ROI Pooling, and no box predictions will be made at that position. 


Based on the above intuition, we conduct the ODA attack by randomly scattering our trigger in the background of the image such that the model can learn an association between the trigger and background. During test time, if the model sees a trigger somewhere in the image, it will believe that it belongs to the background, so that the RPN in FasterRCNN will not predict the object region as an ROI, or the confidence score in the target vector for models of Yolo family will approach 0. If we put the trigger inside an object of any class, the object will disappear since it is regarded as background, which means it is a successful attack under the ODA scenario. The pseudo-code for the scattering process is shown in Algorithm \ref{alg:algorithm}. The intuition is further confirmed by model visualization in Section \ref{main_result_section}.

\begin{algorithm}[tb]
\caption{scatterTrigger()}
\label{alg:algorithm}
\textbf{Input}: $x$: the original image, $T$: the trigger pattern, $\alpha$: the blending ratio, $GT$: the ground truth annotation for the given image, $n$: number of triggers to scatter\\
\textbf{Output}: $x_{poison}$: the poisoned image 

\begin{algorithmic}[1] 
\STATE $x_{poison} \leftarrow x$.copy()
\STATE$k \leftarrow 0$
\STATE$Wt \leftarrow T.width$
\STATE$Ht \leftarrow T.height$
\WHILE{$k < n$}
\STATE$x_{pos} \leftarrow$ randint($(0, x.width)$)
\STATE$y_{pos} \leftarrow$ randint($(0, x.height)$)
\STATE$R \leftarrow$ region\ with\ top\ left\ corner\ $(x_{pos}, y_{pos})$,\\ and bottom\ right\ corner\ $(x_{pos}+Wt, y_{pos}+Ht)$
\STATE$noOverlap \leftarrow true$
\FOR{$i\leftarrow 1$ to $n$}
\IF {IOU($GT[i], R$) $\neq 0$}
\STATE $noOverlap \leftarrow false$                  
\STATE break
\ELSE
\STATE continue
\ENDIF
\ENDFOR
\IF {$noOverlap$}
\STATE $x_{poison}[R] \leftarrow x_{poison}[R] \cdot (1-\alpha) + T \cdot \alpha$
\STATE $GT$.append($R$)
\STATE $k \leftarrow k+1$
\ENDIF

\ENDWHILE
\STATE \textbf{return} $x_{poison}$
\end{algorithmic}
\end{algorithm}

When scattering the triggers in the background, we ensure that there will be no overlapping between any two triggers scattered to maximize the effect of each trigger. This is achieved by line 21 in Algorithm \ref{alg:algorithm}. No trigger shall be scattered into any ground truth bounding boxes to not affect the test time mAP on benign test samples. Possible parameters here include the number of triggers scattered per image, and also the blending ratio, which is an effective tunable parameter to improve stealthiness \cite{chen2017targeted}, between the trigger and the image. The effect of both parameters on our overall attack success rate is investigated in the ablation study under the Experiment Section \ref{ablation_sec}. For easier illustration, an example of the scattering result on an image with the number of triggers scattered per image set as 5, and a blending ratio of 100\%, \textit{i.e.} the original image content is completely replaced, is shown in the examples images in Figure \ref{stage}. However, we do not require such a high number of triggers scattered per image and blending ratio for conducting a successful attack.

\subsection{Object Generation Attack}

\textbf{Object Generation Attack (OGA)} is the scenario where a false positive predicted bounding box of a target class will be generated around the trigger position if a trigger is present at test time. Formally, following the same definition of $Y$ and $F_\theta$ in Section \ref{ODA_meth_sec}, a poisoned test image $x_{poison}$ is constructed at inference by patching $k$ triggers to some random locations $R=\{r_1, ..., r_k\}$ in the background of the image, where $r_i$ denotes a region in the background, satisfying that for $\forall r_i \in R$, $r_i$ is not in $B_j$ for $\forall B_j \in Y$. The goal is that the detection result $Y^{'}$ on the poisoned image will contain false positive bounding boxes at the locations where a trigger is patched, \textit{i.e.} $Y^{'}=F_\theta(x_{poison})=Y \cup \{B_1^{'}, ..., B_k^{'}\}$, where $B_i^{'}$ denotes a false positive bounding box prediction of the target class.

To conduct OGA, we leverage the same step regarding the inherent process of object detectors mentioned in Section \ref{ODA_meth_sec}. Only now instead of looking at the non-objectness regression, we get inspired by the objectness regression:

\begin{equation}
\begin{split}
L_{obj}=\sum_{i\in S\times S} \sum_{j \in B} \mathbb{I}_{ij}^{obj} \cdot (C_{i,j} - \hat{C_{i,j}})^2 
\end{split}
\label{loss2}
\end{equation}
 where $\mathbb{I}_{ij}^{obj}=1$ if anchor box $j$ in cell $i$ is responsible for anchoring an object, and $C_{i,j}$ follows the same definition as in Equation \ref{loss1}. With the same definition of $C$ as in equation \ref{confidence_score}, confidence scores for the grid will be trained to approach the value of $IOU$ if some object of interest lies in this grid since now $Pr(object)$ shall be set as 1. If some confidence scores were higher than the defined threshold in the target vectors, then the model will think that there exists an object of interest in this grid, and the $\mathbb{I}_{i}^{obj}$ value in the classification loss (Equation \ref{loss3}) will be set as 1, making the model penalty wrong classification. 
 \begin{equation}
\begin{split}
L_{cls}=\sum_{i=0}^{S^2} \mathbb{I}_{i}^{obj} \sum_{c \in classes}(p_i(c)-\hat{p_i(c)})^2
\end{split}
\label{loss3}
\end{equation}


Based on the above intuition, we conduct the OGA attack by putting triggers into the center of the ground truth bounding boxes of the target class, so that the model will learn an association between the trigger and the object of the target class. Then the RPN will be more likely to propose the region containing the trigger as an ROI, and the confidence score in the target vector for models of the Yolo family will approach the IOU value between the anchor box and the ground truth. There is only one tunable parameter here in this case, which is the blending ratio. Moreover, there are fewer restrictions during the OGA poisoning process, as putting a trigger into a bounding box is even more trivial than scattering in the background. An example of a poisoned image under OGA can also be found in Figure \ref{stage}. We also further confirm the intuition in Section \ref{main_result_section}.

\section{Experiments}
In this section, we conduct extensive experiments to test the proposed attack on different datasets and models. We first introduce the experiment setup and implementation details in Section \ref{poison_set} and \ref{evl_set}. In Section~\ref{main_result_section}, we show the main results of our baseline setup. We also show model visualizations by showing the corresponding confidence scores. Besides showing the main results of the introduced scenarios, in Section~\ref{ablation_sec}, we conduct ablation studies to comprehensively demonstrate the consistent effectiveness of our method when various variables change. Lastly, we analyze the attack's capability of affecting transfer learning by testing its effectiveness after fine-tuning with clean data in Section \ref{finetuning_subsec}.

\subsection{Poisoning Settings}
\label{poison_set}
\noindent\textbf{Dataset.} We conduct experiments using MSCOCO2017 \cite{cocodataset}, which is a commonly used benchmark dataset for object detection. We use its 118k training images for training and its 5k validation images for evaluation. 

\noindent\textbf{Models.} Our four victim models are Yolov3 \cite{yolov3} with SPP \cite{he2015spatial} structure and a Darknet-53 \cite{darknet13} backbone, Yolov8 \cite{yolov8}, DETR \cite{detr} with ResNet50 \cite{resnet50} feature extractor, and FasterRCNN \cite{fasterRCNN} with ResNet50 backbone. We train Yolov3, Yolov8, DETR, and FasterRCNN using Adam optimizer with a learning rate of 1e-4. This achieves a performance that is close to the best performance that the same models can achieve on the same datasets. 

\noindent\textbf{Poisoning.} We follow previous work on backdoor attacks and define a general poison rate for both the ODA and OGA scenario as the ratio between the number of images we modified and the total number of images in the training dataset. \textit{i.e.} $PoisonRate=\frac{|D_{poison}|}{|D_{train}|}$, where $D_{poison}$ denotes the subset of images that we modify, and $D_{train}$ denotes the full training dataset. For baseline evaluation, we make the following settings: For ODA, The training images are poisoned with a poison rate of 10\% for Yolov3 and Yolov8, and 100\% for DETR and FasterRCNN. We simply poison all objects of the target class for OGA since only a small portion of images contain objects of the target class and this inherently maintains a low poison rate. For OGA, we choose hair dryer as target class with a poison rate of 0.16\%. A patch of interpolated 4$\times$4 Gaussian random noise \cite{hiddenTriggerBackdoor} is chosen as the trigger, as shown in the multimedia appendix. All triggers are blended with a ratio of 100\%. We scatter 5 triggers per image in the ODA scenario.


\subsection{Evaluation Settings} 
\label{evl_set}
\subsubsection{Backdoor Activation}
\label{restriction}
\noindent\textbf{ODA.} 
To evaluate the effectiveness of ODA at inference time, we put a trigger at the center of the target object and check whether the model will fail to detect its existence. To verify the effectiveness of the attack in a more rigorous manner, we set two constraints when attempting to activate the backdoor at inference time:

\textbf{(I)} The original ground truth bounding box is big enough. Concretely, we require $ min(\frac{H}{h_t}, \frac{W}{w_t}) \geq 5 $,
where $H, W, h_t, w_t$ are the height and width of the object bounding box and the trigger respectively. 

\begin{table*}[tp]
    \centering
    \begin{tabular}{c||c|c||c|c||c|c||c|c }
    \hline
   Model & \multicolumn{2}{c||}{Yolov3} & \multicolumn{2}{c||}{Yolov8}  & \multicolumn{2}{c||}{DETR} & \multicolumn{2}{c}{FasterRCNN}\\ \hline
      \makecell{Scenario} & \makecell{ODA} &\makecell{OGA}  &\makecell{ODA} &\makecell{OGA} &\makecell{ODA} &\makecell{OGA} &\makecell{ODA} &\makecell{OGA} \\ \hline
            $ \mathrm{mAP_{normal}} $  & \multicolumn{2}{c||}{44.3} &   \multicolumn{2}{c||}{53.9}  & \multicolumn{2}{c||}{39.9} &  \multicolumn{2}{c}{40.3} \\ \hline
            $ \mathrm{mAP_{benign}} $  & 43.6 & 43.4  & 53.6 & 53.4 &39.1& 39.3& 40.1& 40.2 \\ \hline
             $\mathrm{ASR}$ & 93.5 & 99.1 & 85.9& 96.3&50.7&94.2&77.0&91.9 \\ \hline
              $\mathrm{ASR_{blank}}$ & 6.6 & 0 & 2.5 &0&1.6&0& 2.5 & 0.4\\    
             \hline
        
    \end{tabular} 
    \caption{Main experiment results following the settings in Section \ref{poison_set} and \ref{evl_set}. We report $ \mathrm{mAP_{normal}} $ ,  $ \mathrm{mAP_{benign}} $ , and $\mathrm{ASR}$ (all in \%) on both the ODA and OGA attacking scenarios. Evaluation metrics follow the definitions in Section \ref{sec5.2.2}} 
    \label{main_result}
\end{table*} 

\textbf{(II)} We only put the trigger in the true positive (TP) object predictions, where an object is said to be a TP object if and only if this object is \textbf{\textit{correctly detected and classified}} by the same model when there are no triggers present during inference time, \emph{i.e.}, TP objects are found using the same model on the benign test dataset. Here, the criterion of \textbf{\textit{correctly detected and classified}} is that the bounding box output for this object of interest has an IOU larger than 0.5 with a ground truth bounding box, and the predicted class is consistent with the ground truth label of this object. 

Restriction (I) ensures that the trigger is small enough relative to the object such that the disappearance of an object of interest is not caused by simply covering the object with the trigger, but by the association between the trigger and the background. Restriction (II) ensures that the bounding box of an object disappeared because the backdoor is activated, not simply because the perturbation introduced by the trigger causing the model to miss the object.

\noindent\textbf{OGA.} 
To activate the OGA backdoor at inference time, we put a trigger in the background of the testing image, and check whether a bounding box of the target class is predicted around the trigger location. For more rigorous evaluation, we define the background of a testing image as the regions where no bounding box is predicted by the same model. This prevents miscounting any original existing bounding box predictions as an OGA success.

\subsubsection{Evaluation Metrics}
\label{sec5.2.2}
As the two main goals of the attacker are to make the infected model show comparable performance as a benign model on clean test data, and display the attacker-specified behaviour given the presence of the trigger at test time, we use the mean average precision (mAP) to verify the first goal, and the attack success rate (ASR) to verify the second goal.

Specifically, we define $\mathrm{mAP_{normal}}$ as the mAP that a benign model achieves on a clean test set. We expect this to be close to the best performance that the same model can achieve on the same dataset, so as to ensure the model itself is well converged. We define $\mathrm{mAP_{benign}}$ as the mAP that the infected model achieves on the clean test set. This should be close to $\mathrm{mAP_{normal}}$ so that the infected model shows comparable performance as a benign model on the clean test data. We follow the convention of reporting mAP@0.5:0.95 for the MSCOCO dataset. 


We define ASR for ODA as 
 \begin{equation}
\begin{split}
ASR=\frac{\#\ of\ disappeared\ bbox}{\#\ of\ patched\ trigger\ at\ inference}
\end{split}
\end{equation}
Note that when patching triggers at inference, we follow the restrictions listed in Section \ref{restriction}, which means all disappeared bounding boxes in the numerator must be caused by the activation of the backdoor. This is already guaranteed by the restricted trigger patching process itself.

We define ASR for OGA as 
 \begin{equation}
\begin{split}
ASR=\frac{\#\ of\ generated\ bbox}{\#\ of\ patched\ trigger\ at\ inference}
\end{split}
\end{equation}
Note that the counting of the number of generated bounding boxes is done by iterating through all patched triggers at inference time. This prevents double counting the number of successes of one OGA attack in case more than one bounding box were predicted around the same trigger.  

We also add a set of blank control for comparison to further verify that the backdoor is implanted only after poisoning the training data. Simply adding the trigger on test samples and conducting inference with a clean model will not lead to a successful attack. The attack success rate of a clean model on the poisoned test set is denoted as $\mathrm{ASR_{blank}}$.



\subsection{Main Results}
\label{main_result_section}
Following the poisoning and evaluation settings defined in Section \ref{poison_set} and \ref{evl_set}, the main results are shown in Table \ref{main_result}. It can be seen that the model achieves $\mathrm{mAP_{normal}}$ that is close to the best performance that the same model can achieve on the MSCOCO dataset, suggesting the model itself is well converged under our training settings. With the same training settings, the infected model achieves an $\mathrm{mAP_{benign}}$ that is very close to the value of $\mathrm{mAP_{nomral}}$, showing that the infected model behaves very similarly to a clean model on the clean testing data. Finally, we achieve an ASR of more than 90\% on both scenarios using Yolov3, an ASR of 85\%+ and 95\%+ on ODA and OGA respectively using Yolov8, an ASR of more than one half for ODA and more than 90\% for OGA using DETR, and an ASR of 77\% and 91.9\% on ODA and OGA respectively using FasterRCNN. These ASR values are threatening high for safety-critical applications, proving the overall effectiveness of the proposed attack.

\begin{table*}[tp]
\setlength\tabcolsep{1.05pt}
\renewcommand{\arraystretch}{1}
\centering
\begin{tabular}{@{}c|ccc |ccc |ccc |ccc |ccc|ccc@{}}
\hline

 &  \makecell{Trigger\\Size}   & ASR & mAP &  \makecell{Trigger\\Pattern} & ASR & mAP &   \makecell{Blended\\Ratio}   & ASR & mAP  &   \makecell{Trigger\\Number}   & ASR & mAP &   \makecell{Poison\\Rate}   & ASR & mAP &\makecell{Target\\Class}   & ASR & mAP  \\ \hline
 {\multirow{6}{*}{ODA}}  & \multirow{2}{*}{30$\times$30}  & \multirow{2}{*}{99.5} &  \multirow{2}{*}{43.5}   &   \multirow{2}{*}{Noise} & \multirow{2}{*}{99.5} &   \multirow{2}{*}{43.5} & \multirow{2}{*}{100$\%$}  & \multirow{2}{*}{99.5} & \multirow{2}{*}{43.4} &  \multirow{2}{*}{5}  & \multirow{2}{*}{99.5} & \multirow{2}{*}{43.4} & 100$\%$ & 99.5 & 43.5  &--&--&--  \\
                &  &  &  &  &  &  & &  &  &  &  &  & 50$\%$ & 93.1 & 43.5 &--&--&--\\
                             & \multirow{2}{*}{20$\times$20}  & \multirow{2}{*}{97.7} &  \multirow{2}{*}{43.2}  &   \multirow{2}{*}{C-Mark} & \multirow{2}{*}{99.4} &   \multirow{2}{*}{43.6} & \multirow{2}{*}{80$\%$}  & \multirow{2}{*}{98.3} & \multirow{2}{*}{42.6} & \multirow{2}{*}{2}  & \multirow{2}{*}{98.8} & \multirow{2}{*}{43.0} & 20$\%$& 91.9& 42.0 &--&--&--
                                  \\
                 &  &  &  &  &  &  & &  &  &  &  &  & 10$\%$ & 93.5 & 43.6 &--&--&--\\
                     & \multirow{2}{*}{10$\times$10}  & \multirow{2}{*}{97.9} & \multirow{2}{*}{42.1} &   \multirow{2}{*}{Melon} & \multirow{2}{*}{96.8} &   \multirow{2}{*}{42.4} & \multirow{2}{*}{50$\%$}  & \multirow{2}{*}{98.7} & \multirow{2}{*}{42.3}  & \multirow{2}{*}{1}  & \multirow{2}{*}{99.3} & \multirow{2}{*}{42.7} & 5$\%$& 92.6 & 43.9 &--&--&--    \\
                &  &  &  &  &  &  & &  &  &  &  &  & 1$\%$ & 43.9 & 42.5 &--&--&-- \\
                            \hline
 {\multirow{3}{*}{OGA}}     & 30$\times$30 & 99.1 & 43.4 & Noise & 99.1 & 43.4 & 100$\%$ & 99.1 & 43.4 &--&--&--&--&--&--& \makecell{HairDrier} & 99.1 & 43.4 \\
                              & 20$\times$20 & 97.7 & 42.1 & C-Mark & 93.5 & 42.1 &   80$\%$ & 99.7 & 42.8 &--&--&--&--&--&--& Toaster & 98.3 & 42.8  \\
                              & 10$\times$10 & 86.0 & 43.7 & Melon & 95.1 & 42.9 & 50$\%$ & 95.3 & 43.1 &--&--&--&--&--&--& Scissors & 99.6 & 43.2  \\

  \hline
\end{tabular}
\caption{Results of ablation studies. We show the influence of each variable in the attack settings under both scenarios. The ASR here follows the same definition under main results, and the mAP refers to $\mathrm{mAP_{benign}}$ as defined in evaluation settings.}
\label{ablation_table}

\end{table*}

To further help us understand how the association between the trigger and the background is built throughout the ODA training process and how the association between the trigger and objects of the target class is built throughout the OGA training process, we also inspect the backdoor implanting process by making an evaluation of the ASR after every epoch of training. Details are listed in Appendix A.

Despite that the ASR value already proves the effectiveness of the attack, we confirm the intuition of our method by ``seeing inside" the model. We plot heat-maps visualizing the magnitude of the confidence scores for Yolo and the magnitude of the RPN foreground score for FasterRCNN on both the clean and infected models under both the ODA and OGA scenarios. The visualizations are shown in Appendix B. The heat map visualizations further confirm our intuition of the attacking strategy.

\begin{figure}[!h]
\centering
\includegraphics[width=0.2\textwidth]{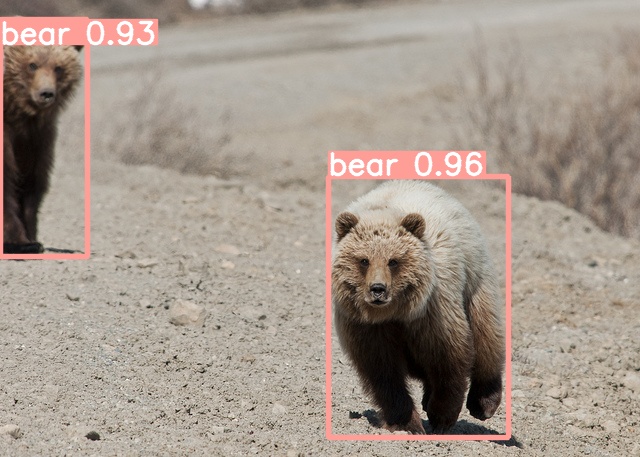}
\includegraphics[width=0.2\textwidth]{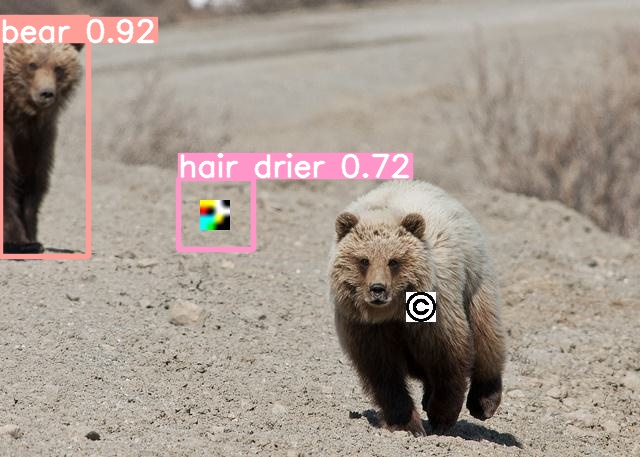}
\caption{Example of simultaneous ODA and OGA. Image on the left is the original image. Image on the right is patched with the Gaussian noise for OGA and the Copyright watermark for ODA.}
\label{ODA_OGA_simultaneous}
\end{figure}

\noindent\textbf{Simultaneous OGA and ODA.} We further explored an interesting scenario of conducting ODA and OGA simultaneously by setting one trigger pattern for ODA and another pattern for OGA. Results show that they can co-exist while achieving good individual attacking performance. (ASR \textbf{92.0\%} and \textbf{98.7\%} for ODA and OGA respectively). An visualization example of this scenario is shown in figure \ref{ODA_OGA_simultaneous}.

\noindent\textbf{Generalizability.} Generalizability is an important property for a realistic attack. To test whether the implanted backdoor can be activated when the infected model is deployed to real world images, we deploy the model on a surrogate dataset to simulate large scale generalization. By deploying the COCO trained infected model on BDD100K \cite{bdd100k}, we achieve ASR \textbf{97.5\%} and \textbf{91.2\%} for ODA and OGA respectively. More visualizations on random real world images are shown in the multimedia appendix.

\begin{figure}[!h]
    \centering
    \includegraphics[width=0.23\textwidth]{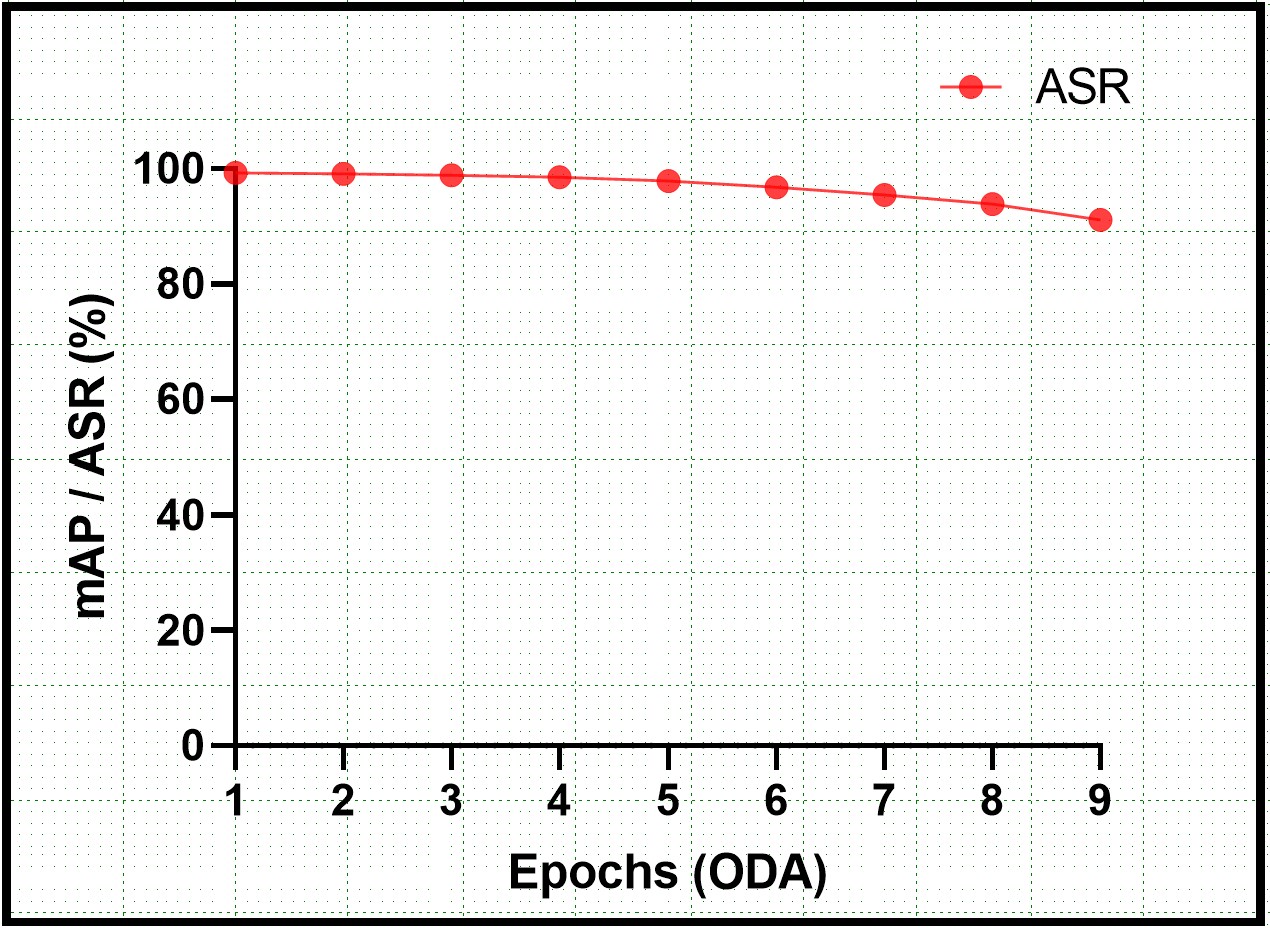}
    \includegraphics[width=0.23\textwidth]{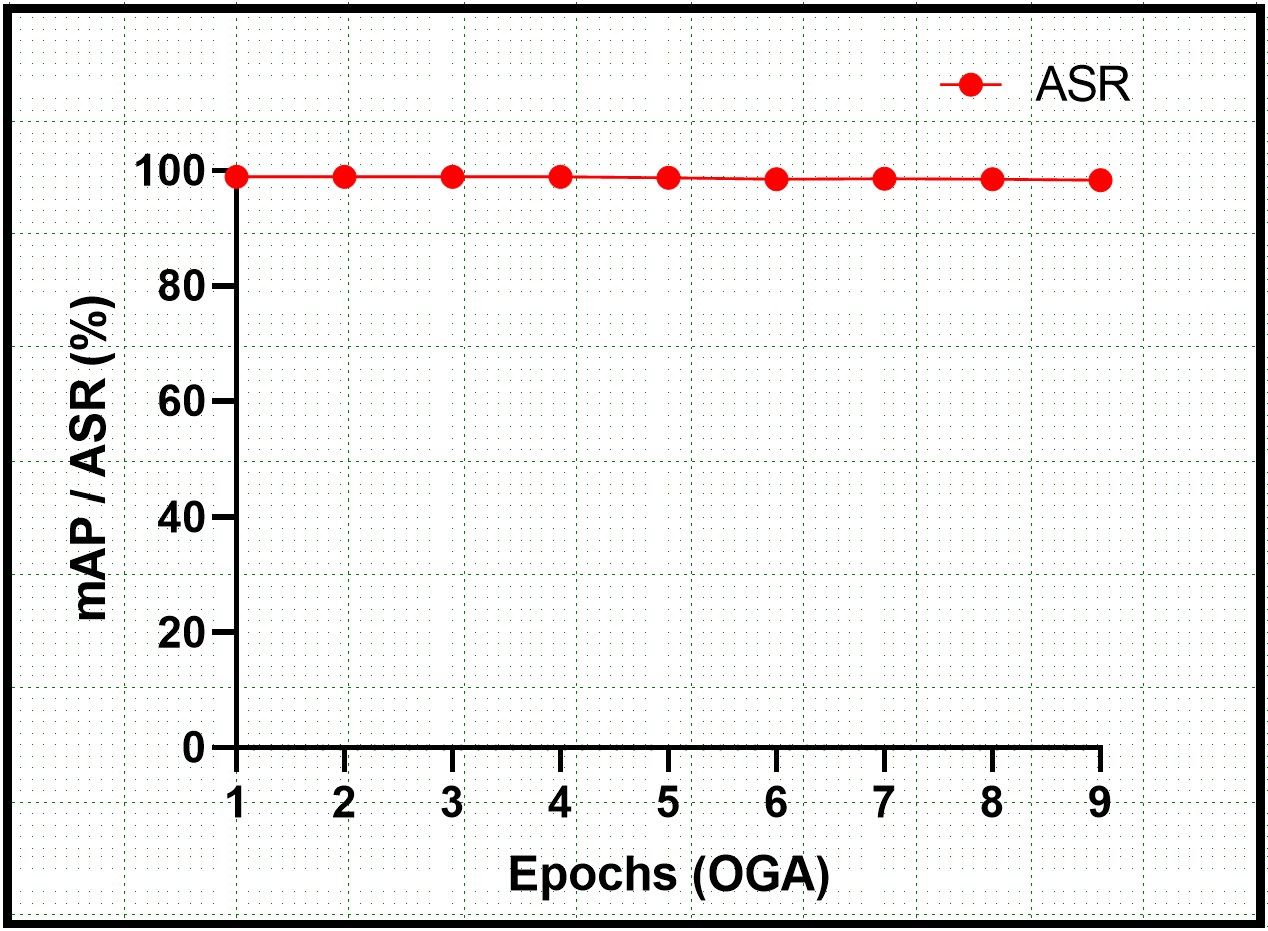}
   \caption{The result of fine-tuning the infected model on a clean PASCAL VOC07+12 training set + a clean MSCOCO validation set, which is 21k clean images in total.}
    \label{finetuning}
\end{figure}

\subsection{Ablation Study}
\label{ablation_study_section}
To explore how each variable in our settings may alter our attack effectiveness, we conduct ablation studies to investigate the influence of each parameter, which include the trigger size, trigger pattern (shown in the multimedia appenidx), trigger blending ratio under both ODA and OGA, target class selection under OGA, and the number of triggers scattered per image and the poison rate under ODA. The complete results are shown in Table \ref{ablation_table}. When investigating the influence of one variable, we control all other variables by setting default values as the followings: We set the number of scattered triggers per image for OGA as 5, the poison rate for ODA as 100\%, trigger size as 30$\times$30, blending ratio as 100\%, trigger pattern as the patched 4$\times$4 Gaussian noise, and the target class for OGA as hairdryer. We can see that the attack works effectively under different variable settings.

\label{ablation_sec}



 


      

\subsection{Sustainability under Fine-tuning}
\label{finetuning_subsec}

Transfer learning is commonly adopted in training neural networks as people may often download pre-trained models from a third party and fine-tune it on a custom dataset. Fine-tuning, which can be easily applied to the object detection setting, has been proven to be an effective way of mitigating backdoors in neural networks \cite{sha2022fine}. We fine-tune our infected model, which is trained on a poisoned MSCOCO training set, using a clean PASCAL VOC07+12 \cite{pascal-voc-2007,pascal-voc-2012} training set + a clean MSCOCO validation set, which is 21K clean images in total. The fine-tuning results are shown in Figure \ref{finetuning}. It can be seen that the attack success rate of OGA still remains nearly 100\% after fine-tuning, and although the ASR of ODA slightly dropped, it still maintains a value of more than 90\%. This proves that our attack can also be used against transfer learning.

\section{Conclusion}
In this paper, we revealed an inherent property of deep learning-based object detectors, which can be leveraged to conduct backdoor attacks. Based on this inherent property, we proposed a simple, intuitive, yet effective backdoor attack poisoning strategy against object detection applications in a clean-labeled manner under the ODA and OGA scenario. Specifically, without modifying annotations, we design the poisoning strategy by aligning with the association built by the model itself. Extensive experiments verify the effectiveness and generalizability of our attack under different settings, which also show that our attack can induce threats towards transfer learning as fine-tuning is unable to mitigate the implanted backdoor. We hope this work can raise the community's awareness of the potential threat of backdoor attacks to object detectors, which are extensively used in a wide range of safety-sensitive real-life applications.


\bibliography{aaai24}


\begin{appendices}

\section{Backdoor Implanting Process}
\begin{figure}[!h]
    \centering
    \includegraphics[width=0.23\textwidth]{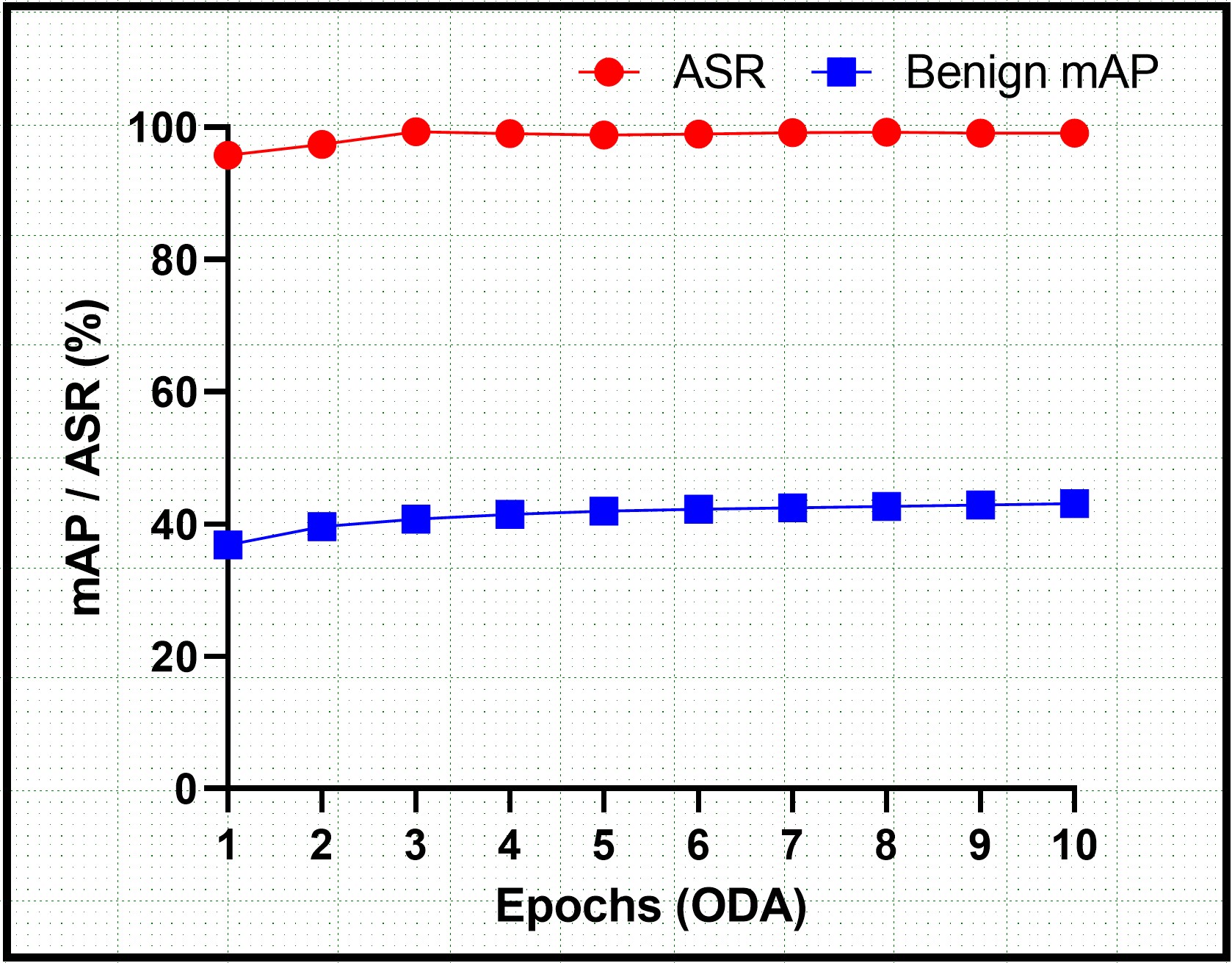}
    \includegraphics[width=0.23\textwidth]{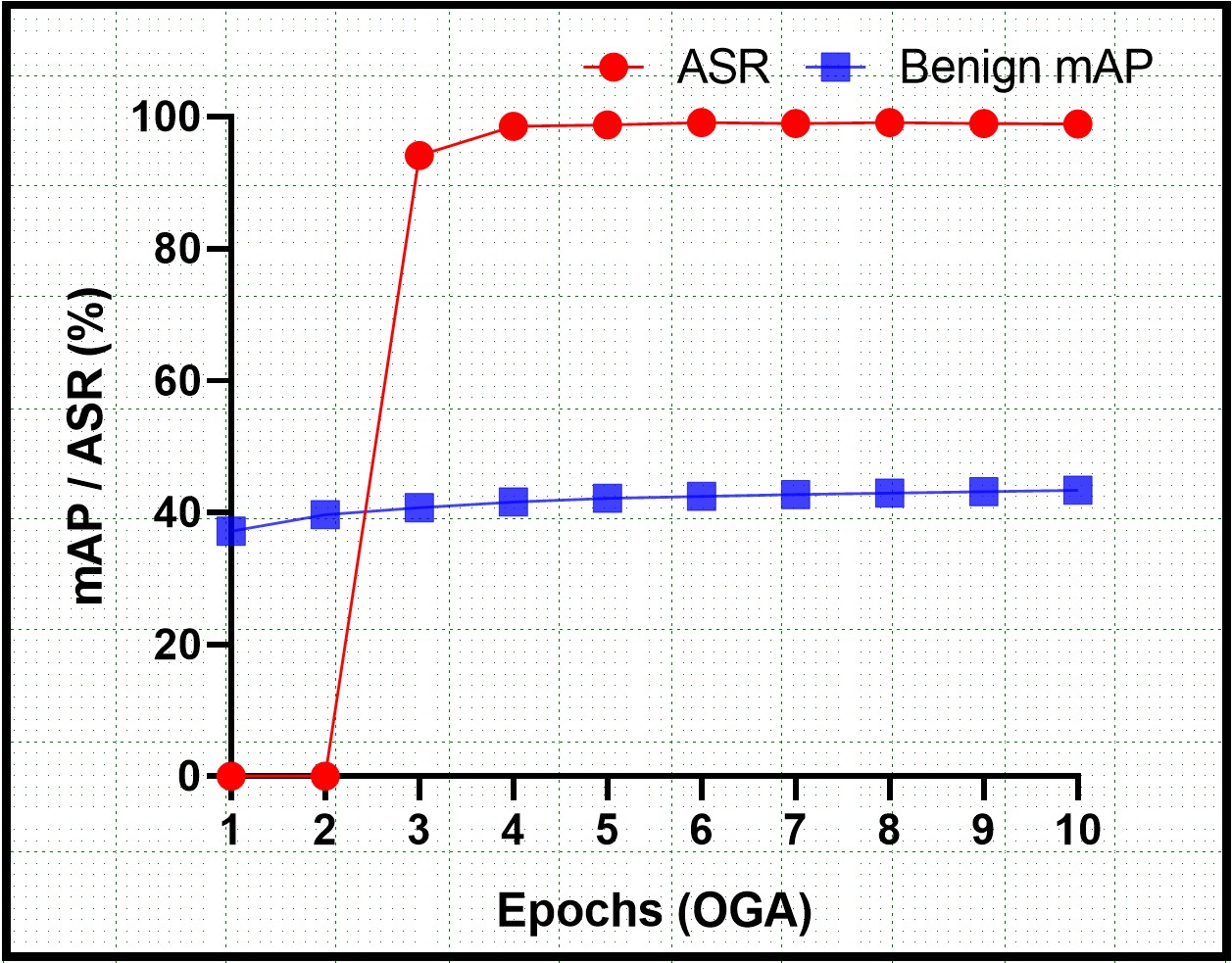}
    
   \caption{The plot showing the changes of $mAP$ and $ASR$ of Yolov3 with the training epochs on both ODA and OGA.}
    \label{asr_map_in_training}
\end{figure}

To further help us understand how the association between the trigger and the background is built throughout the ODA training process and how the association between the trigger and objects of the target class is built throughout the OGA training process, we make an evaluation of the ASR after every epoch of training. The plot showing the association built-up progress is shown in Figure \ref{asr_map_in_training}. From the plot, we can see that both the $mAP$ and $ASR$ are gradually growing as more epochs are trained, proving that the desired associations are indeed built throughout the training process. We also discover that in the ODA scenario, the ASR is already very high just after one epoch of training, but it takes slightly longer for the ASR to grow under the OGA scenario. We suspect that it is more challenging for the model to build an association between the trigger and a specific class than building the association with the background. This may be caused by the fact that similar to the feature of an effective trigger, which should be something easy to learn \cite{sandoval2022poisons}, many features of background areas are similar to low-level textures, such as brick walls, plain sky, calm water, etc., which are also easy features for the neural network. Hence it is easier for the model to regard the trigger and the background as the same thing. However, features of the specific classes contain much more high-level abstracts, making it harder for the model to find the similarity between the trigger and the class.
\section{Model Visualization}
Despite that the ASR value already proves the effectiveness of the attack, we confirm the intuition of our method by ``seeing inside" the model. We take the output of the (40$\times$40) header of the Yolov3 model and the output of the RPN in FasterRCNN as examples. 

\begin{figure*}[tp]
    \centering
    \includegraphics[width=0.49\textwidth]{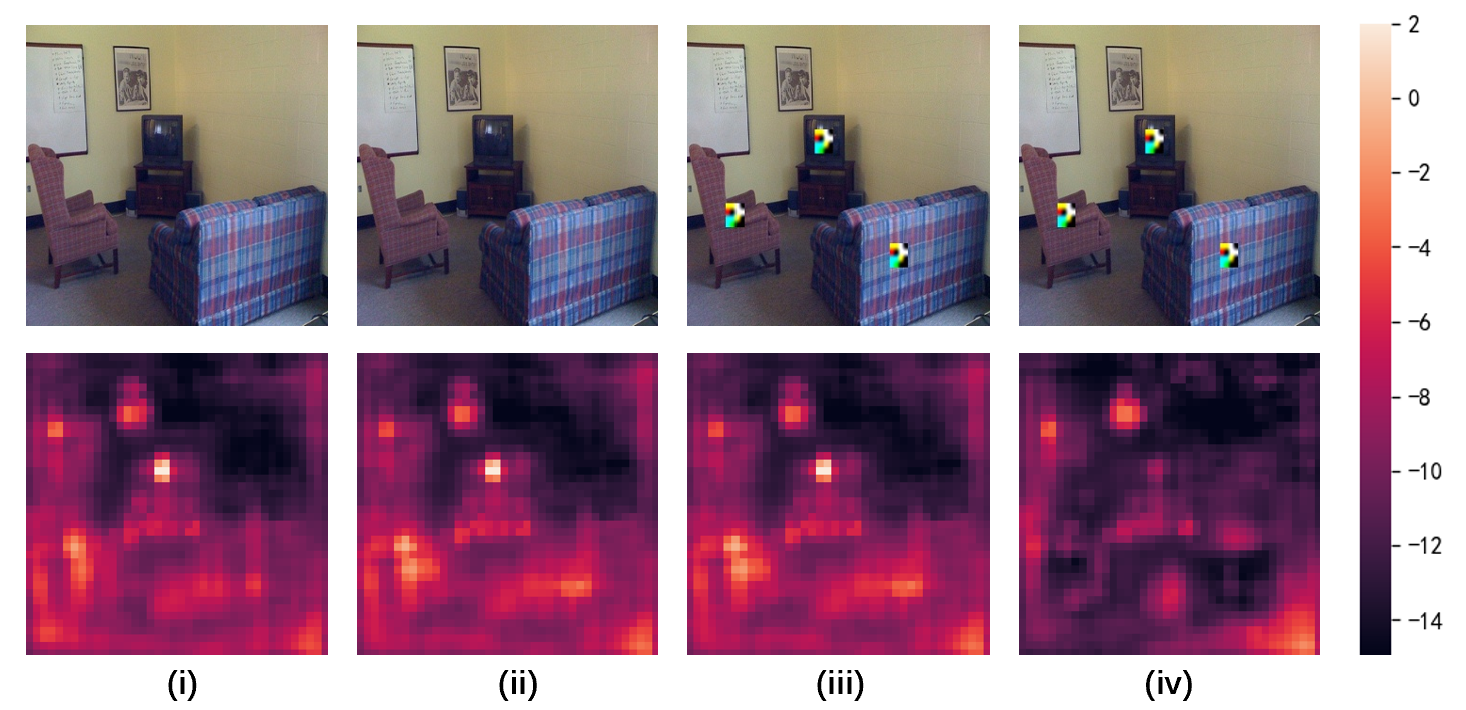}
    \includegraphics[width=0.49\textwidth]{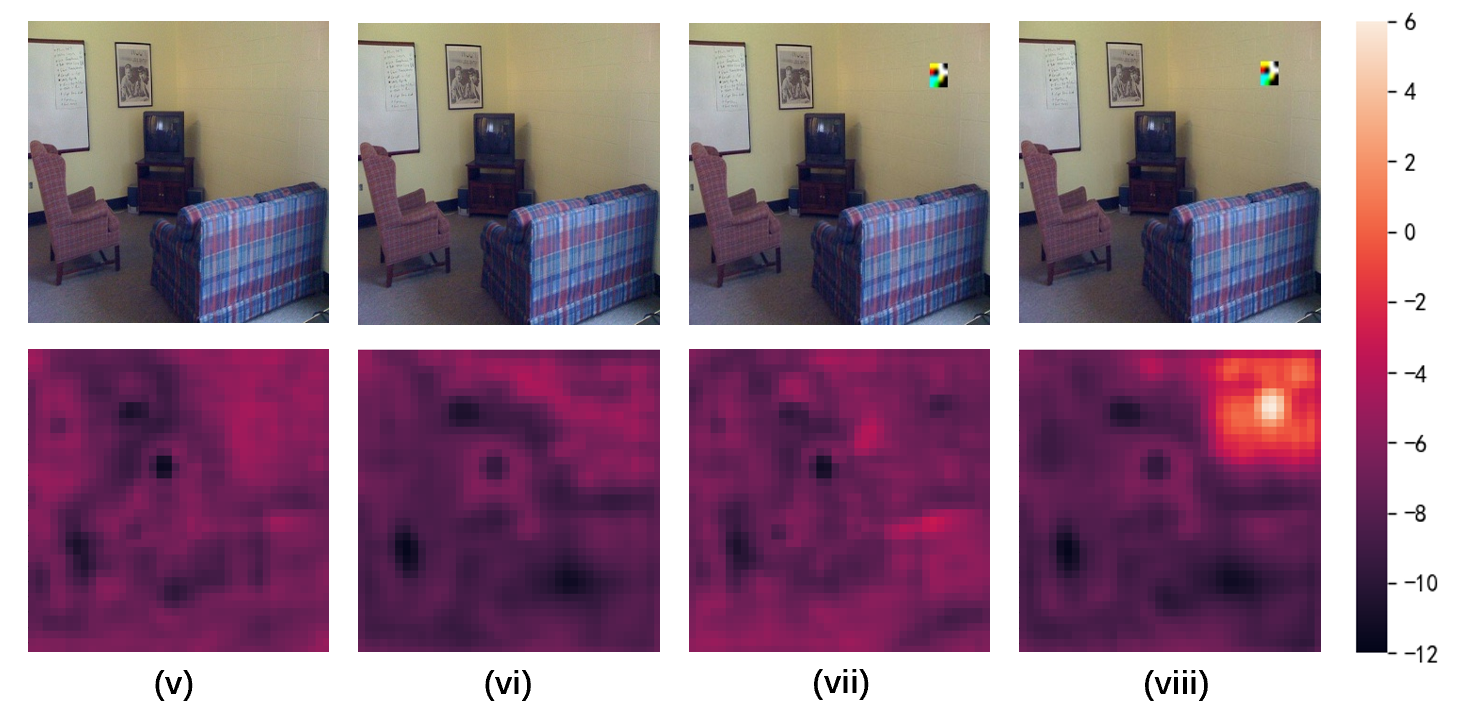}
   \caption{Heat maps for model visualization to confirm our intuition. The Yolov3 \cite{yolov3} model has three prediction headers of size (40$\times$40), (80$\times$80), and (160$\times$160) each. We choose the (40$\times$40) header as an example. We show the objectness confidence score for ODA and the target class confidence score for OGA. Sub-figure (i)-(iv) are the ODA prediction result of the clean model on clean data, the infected model on clean data, the clean model on poisoned data, and the infected model on poisoned data respectively, and (v)-(viii) are the OGA prediction results under the same order of model-data combinations.} 
    \label{heatmap}
\end{figure*}

For Yolov3, we plot a heat map illustrating the magnitude of the confidence scores on both a clean model and an infected model under both the ODA and OGA scenarios. The heat map is illustrated in Figure \ref{heatmap}. The shade of the color in the heat map represents the confidence score. It can be seen that under the ODA scenario, the infected model indeed produced a significantly lower object confidence score when a trigger is patched to an object at inference (Figure \ref{heatmap}.(iv)), and produced a significantly higher specific-class confidence score when a trigger is patched to the background area during inference under the OGA setting (Figure \ref{heatmap}.(viii)). 

\begin{figure*}[tp]
    \centering
    \includegraphics[width=0.49\textwidth]{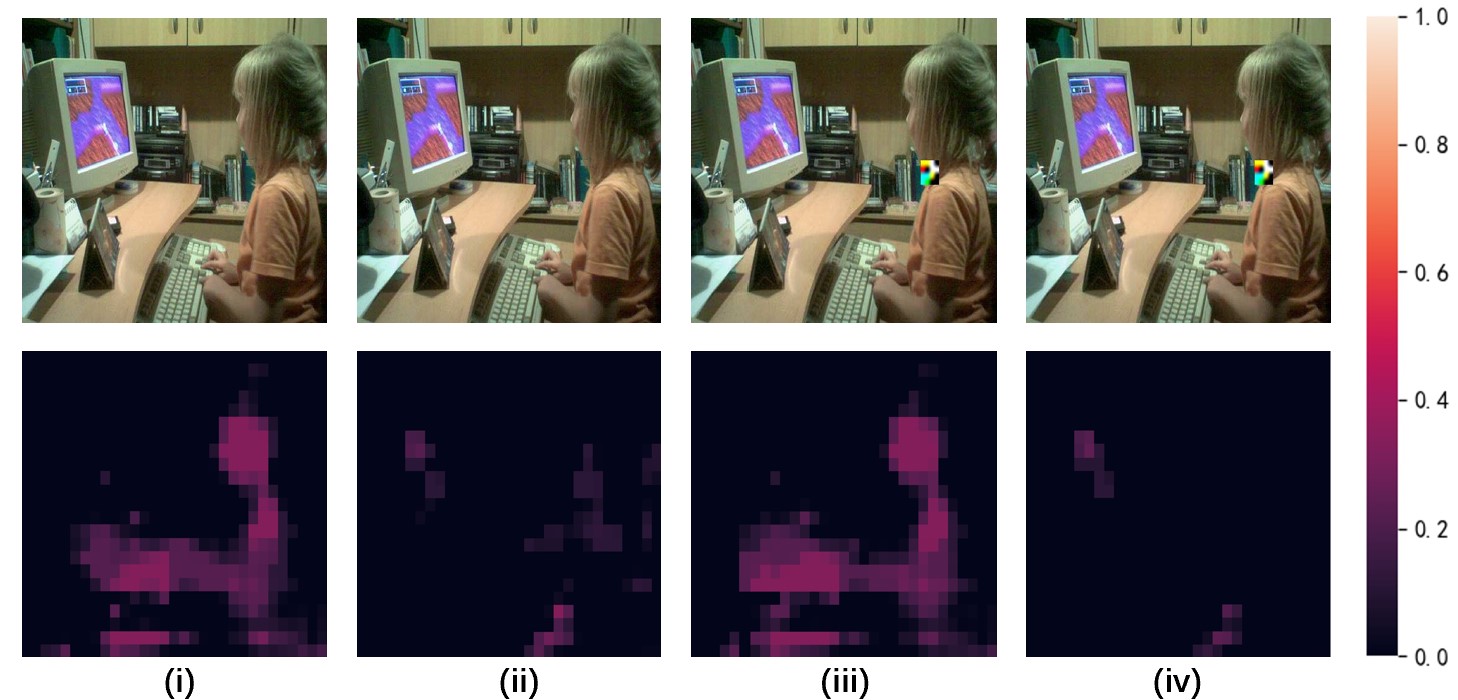}
    \includegraphics[width=0.49\textwidth]{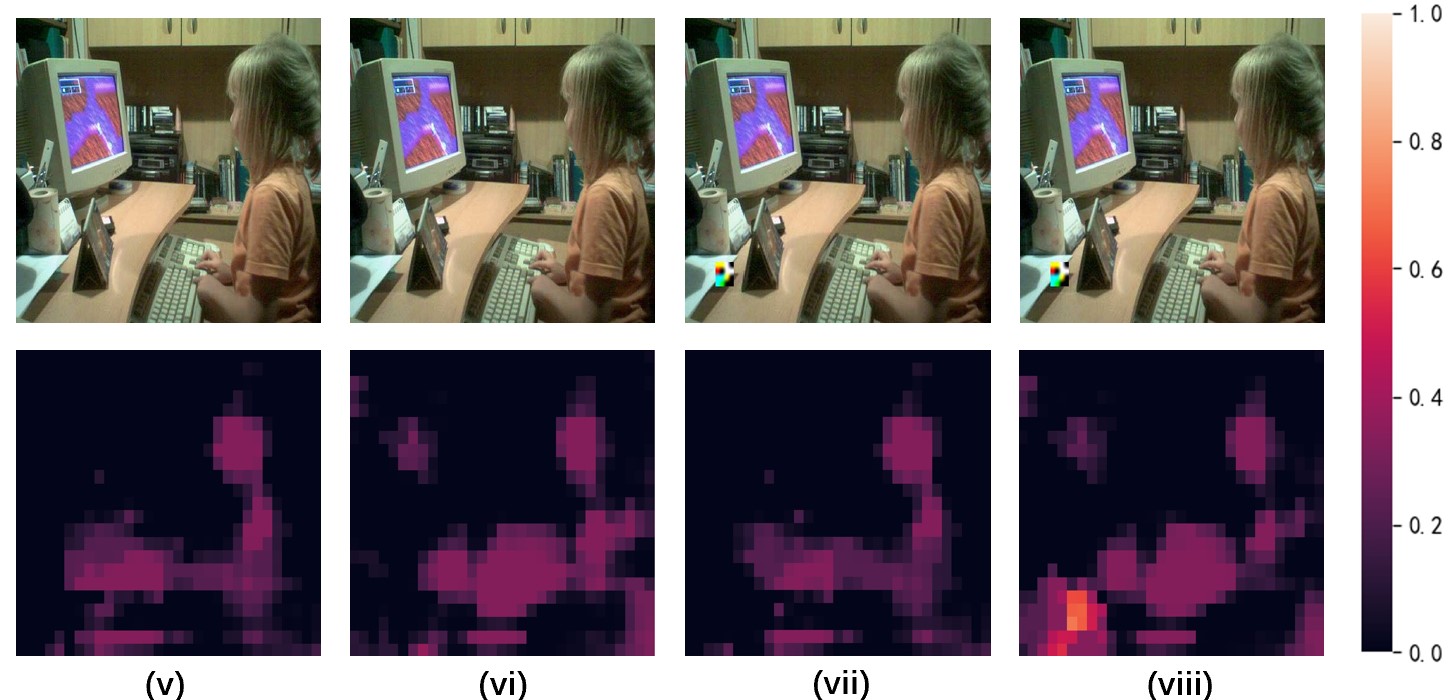}
   \caption{Heat maps for model visualization of FasterRCNN. We show the RPN foreground score for ODA and OGA. Sub-figure (i)-(iv) are the ODA prediction result of the clean model on clean data, the infected model on clean data, the clean model on poisoned data, and the infected model on poisoned data respectively, and (v)-(viii) are the OGA prediction results under the same order of model-data combinations.} 
    \label{frcnn_heatmap}
\end{figure*}

For FasterRCNN, we plot a heat map illustrating the magnitude of the RPN foreground scores on both a clean model and an infected model under both the ODA and OGA scenarios. The heat map is illustrated in Figure \ref{frcnn_heatmap}. The shade of the color in the heat map represents the RPN foreground score. Similarly in Yolov3, it can be seen that under the ODA scenario, the infected model indeed produced a significantly lower  RPN foreground score when a trigger is patched to an object at inference (Figure \ref{frcnn_heatmap}.(iv)), and produced a significantly higher RPN foreground score when a trigger is patched to the background area during inference under the OGA setting (Figure \ref{frcnn_heatmap}.(viii)).These heat map visualizations further confirm our intuition of the attacking strategy.

\end{appendices}

\end{document}